\pdfoutput=1
\documentclass{article}

\usepackage{arxiv}

\usepackage[version=3]{mhchem} 
\usepackage{amsfonts}
\usepackage{pgfplots}
\pgfplotsset{compat=1.18}
\usetikzlibrary{decorations.pathreplacing}
\usepackage{subfig}
\usepackage{url}
\usepackage{hyperref}

\begin{document}
\title{Reversible Deep Learning for $^{13}$C NMR in Chemoinformatics: On Structures and Spectra}

\author{ \href{https://orcid.org/0000-0002-5990-4157}{\includegraphics[scale=0.06]{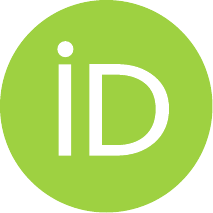}\hspace{1mm}\textbf{Stefan Kuhn}} \\
Department of Computer Science\\
Tartu University, Tartu, Estonia\\
and De Montfort University, Leicester, UK \\
\texttt{stefan.kuhn@dmu.ac.uk}\\
\and \href{https://orcid.org/0000-0001-7852-4961}{\includegraphics[scale=0.06]{orcid.pdf}\hspace{1mm}\textbf{Vandana Dwarka}}\\
Department of Industrial and Applied Mathematics\\
Delft University of Technology\\
Delft, the Netherlands\\
\texttt{v.n.s.r.dwarka@tudelft.nl}
\and  \href{https://orcid.org/0009-0000-5944-769X}{\includegraphics[scale=0.06]{orcid.pdf}\hspace{1mm}\textbf{Przemyslaw Karol Grenda}}\\
Faculty of Pharmacy\\
Charles Universisty\\
Hradec Králové, Czech Republic\\
\texttt{grendap@faf.cuni.cz}
\and  \href{https://orcid.org/0000-0003-1902-9383}{\includegraphics[scale=0.06]{orcid.pdf}\hspace{1mm}\textbf{Eero Vainikko}}\\
Department of Computer Science\\
Tartu University\\
Tartu, Estonia\\
\texttt{eero.vainikko@ut.ee}
}

\maketitle              
%

\begin{abstract}
We introduce a reversible deep learning model for $^{13}$C Nuclear Magnetic Resonance (NMR) spectroscopy that uses a single conditional invertible neural network for both directions between molecular structures and spectra. The network is built from i-RevNet style bijective blocks, so the forward map and its inverse are available by construction. We train the model to predict a 128-bit binned spectrum code from a graph-based structure encoding, while the remaining latent dimensions capture residual variability. At inference time, we invert the same trained network to generate structure candidates from a spectrum code, which explicitly represents the one-to-many nature of spectrum-to-structure inference. On a filtered subset, the model is numerically invertible on trained examples, achieves spectrum-code prediction above chance, and produces coarse but meaningful structural signals when inverted on validation spectra. These results demonstrate that invertible architectures can unify spectrum prediction and uncertainty-aware candidate generation within one end-to-end model. 
\end{abstract}


\section{Introduction}

\begin{figure}
  \centering
  \subfloat[A three dimensional representation (generated with PyMOL\cite{PyMOL}).]{%
    \includegraphics[width=0.48\textwidth]{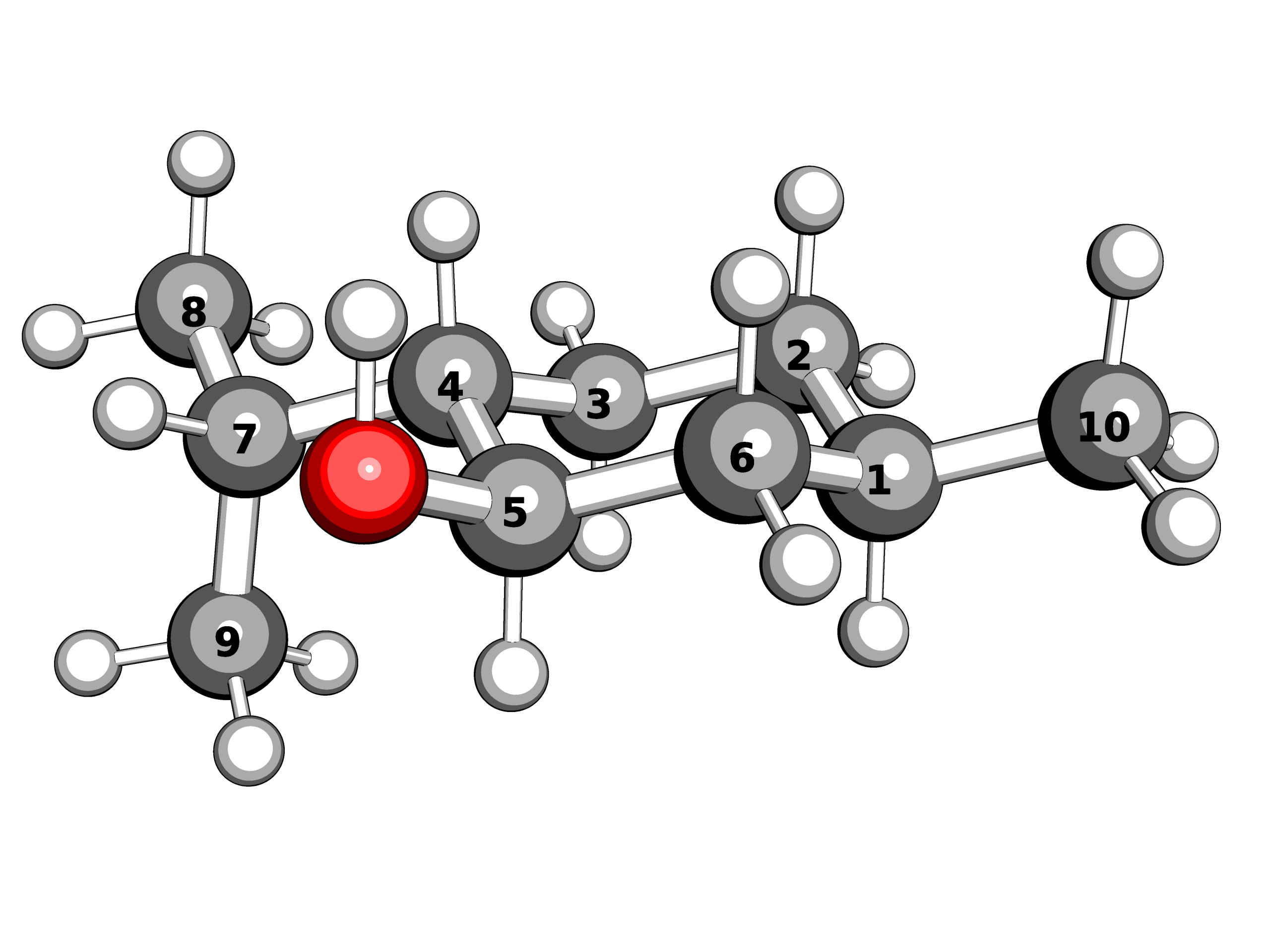}%
  }\hfill  
  \subfloat[A representation as a 2D graph (generated by nmrshiftdb.org).]{%
    \includegraphics[width=0.48\textwidth]{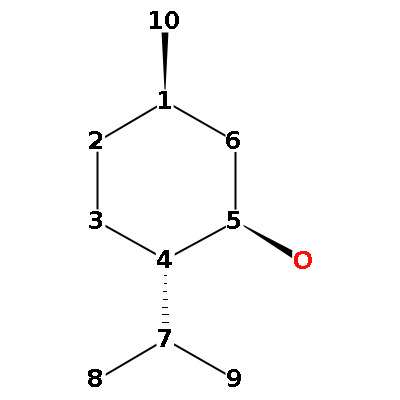}%
  }
  \caption{Two ways to represent the structure of (-)-Menthol.}
  \label{fig:menthol}
\end{figure}

Chemistry, according to the Encyclopaedia Britannica\footnote{\url{https://www.britannica.com/science/chemistry}, accessed 20.1.2026}, ``deals with the properties, composition, and structure of substances (defined as elements and compounds), the transformations they undergo, and the energy that is released or absorbed during these processes.''. For doing so, one important component is the modeling of entities, for our purposes, we focus on molecules (which we can consider to be the same as compounds). Figure~\ref{fig:menthol}~a) shows a ``realistic'' model of a molecule, whereas Figure~\ref{fig:menthol}~b) displays a highly stylized depiction, where atoms and bonds form a graph. Whilst both are not fully realistic (e. g. atoms do not show colours), Figure~\ref{fig:menthol}~b) is clearly even more abstract. Still, such models form a backbone of chemical research until today.

\begin{figure}
  \centering
  \includegraphics[width=1\textwidth]{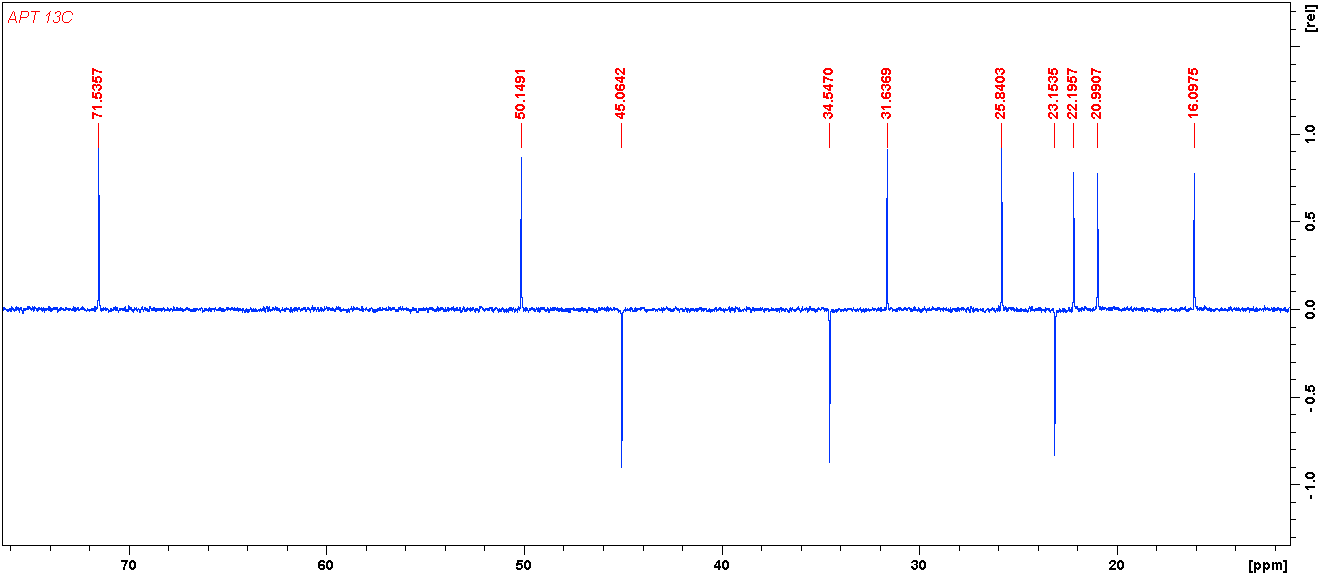}%
  \caption{A $^{13}C$ spectrum of (-)-Menthol, measured at 150 MHz in CDCl3 \cite{berger2009classics}.}
  \label{fig:spectrum}
\end{figure}

Another important field in chemistry is how to identify molecules and how to determine their structures. Since molecules are invisible, various methods from the field of analytical chemistry need to be employed here. One of the most powerful methods is nuclear magnetic resonance (NMR) spectroscopy. There are a multitude of experiments in NMR, we concentrate on one of the most popular and relatively simple from the NMR family, $^{13}C$ NMR spectroscopy. The result of performing this NMR experiment on the structure in Figure~\ref{fig:menthol}, is shown in Figure~\ref{fig:spectrum}. The details of such a spectrum are not relevant for this paper, it is sufficient to say that peak frequency (the X coordinate) is related to the specific atom's chemistry within the molecule, with molecular symmetry acting as a dimensionality reduction factor. Since the position of a peak on the X-axis depends on the local chemical environment of the atom, and those are different for different atoms, the peaks form a characteristic fingerprint that strongly constrains molecular identity. Similar to the molecule, the spectrum can be simplified, in the simplest case to a list of X-values in ppm. In our case this would be [31.6, 34.6, 23.2, 50.2, 71.5, 45.1, 25.8, 21, 16.1, 22.2], where the ten values correspond to the ten signals of the atoms in the molecule (ordered by the atom numbers in Figure~\ref{fig:menthol}).

Since the relationship between molecules and spectra is ultimately ruled by laws of nature, the relationship can be and has been modeled. This is possible and relevant in both directions: On the one hand, it is a common task in chemoinformatics to find the peak values for a compound. This exercise, known as \textit{spectrum prediction} can be useful, for example, if a chemist knows that an unknown structure can only be one of a few candidates. By measuring its spectrum and comparing it to the predicted spectra for the candidates, it is possible to identify the correct candidate. The target of a prediction model is to predict the peak values as accurately as possible and to minimize the error. Modern techniques based typically on graph neural networks can achieve very low errors. We refer the reader to recent reviews \cite{Das2025,PMID:34787335}.

On the other hand, the reverse exercise is important as well: A chemist isolates a substance, measures its spectrum, and a computer program tells which molecule is behind this. Such programs are known as \textit{Computer-Aided Structure Elucidation} (CASE) systems. Traditional CASE software uses optimization techniques to find the best match in the (very large) chemical space. A recent example is \cite{molecules28031448}. Very recently, there were attempts at training a neural network to do this task directly without optimization, e. g. \cite{hu2025pushinglimitsonedimensionalnmr}. Clearly, advanced network architectures are needed for this.



Learning a reversible mapping between molecular structures and NMR spectra naturally connects to research on invertible neural architectures and probabilistic inverse problems. In $^{13}C$ NMR, the forward direction (structure → spectrum) is typically well defined, while the inverse direction (spectrum → structure) is inherently ambiguous: multiple molecular structures, conformations, or environments may give rise to similar spectral signatures.  This asymmetry motivates models that preserve information in the forward direction while explicitly representing uncertainty in the inverse direction \cite{Ardizzone2021cINN}.

Invertible and reversible neural networks challenge the conventional paradigm of progressively compressive feature extraction. The i-RevNet architecture demonstrates that deep networks can be constructed entirely from bijective blocks, preserving all information throughout the network depth and enabling exact inversion of intermediate representations \cite{DBLP:journals/corr/abs-1802-07088}, forming an \textit{Invertible Neural Network} (INN). This idea is closely related to reversible residual networks, where paired residual updates allow the input to be reconstructed without storing activations, offering both theoretical and practical benefits \cite{Gomez2017RevNet,10.1007/978-3-031-97063-4_12}. Such architectures show that information preservation is compatible with deep hierarchical representations and provide a natural basis for bidirectional mappings between physical parameters and observations.

Reversible modeling has also been explored in recurrent settings, where it was shown that strict reversibility limits implicit “forgetting.” To remain expressive, reversible models must explicitly track information that would otherwise be discarded \cite{10.5555/3327546.3327578}. This insight is particularly relevant for spectroscopy: uncertainty in the inverse mapping cannot be eliminated by architectural depth alone and must instead be represented explicitly, for example via latent variables or conditioning mechanisms.

INNs and normalizing flows provide a probabilistic framework well suited for such inverse problems. By learning a bijective transformation between data and a latent space with a simple base distribution, INNs allow exact likelihood evaluation and efficient sampling. Canonical flow-based constructions such as NICE, RealNVP, and Glow introduced coupling layers and invertible convolutions that scale invertible models to high-dimensional data \cite{Dinh2014NICE,Dinh2016RealNVP,Kingma2018Glow}. In the context of inverse problems, INNs are typically augmented with additional latent dimensions to encode information that cannot be recovered uniquely from the measurements. Sampling these latent variables yields a distribution over solutions consistent with the observed data, rather than a single point estimate \cite{Ardizzone2018InverseINN}.

Conditional INNs extend this framework by separating the roles of invertibility and conditioning. The core bijective transformation remains invertible, while an auxiliary conditioning network extracts features from the observation (e.g., a spectrum) and injects them into the coupling layers \cite{Ardizzone2021cINN}. This design enables one-to-many inverse mappings while maintaining a consistent forward model. In practice, the conditioning path can incorporate domain-specific encoders for spectral data, while the invertible path enforces structural consistency between molecular representations and spectra.

Because NMR modeling is often discussed alongside autoencoders, it is important to clarify the distinction. Standard autoencoders learn separate encoder and decoder mappings that are not mutually inverse and typically rely on a lossy bottleneck. While variational autoencoders introduce a probabilistic latent space, reconstruction and invertibility remain approximate. In contrast, INNs enforce an exact bijection between input and latent variables (up to intentional dimensional augmentation), fundamentally changing how information loss and uncertainty are handled. Recent work explicitly comparing autoencoders and INNs shows that INN-based formulations can preserve information while still supporting dimensionality reduction through structured latent variables \cite{Nguyen2023INNAE}.



Considering this, the molecule-$^{13}C$ spectrum relationship is an ideal candidate for applications of invertible neural networks. We have very information-rich structures on both sides, the information is different, but there is a strict relationship between them. So there is a reversible, invertible connection between molecular structures and NMR spectra that preserves forward consistency while explicitly modeling inverse ambiguity. That relationship has been shown as learnable in both directions. Therefore, in this paper, we present the first invertible neural network, which delivers an approximation of spectrum simulation in one direction and of structure elucidation in the other direction. To our knowledge, this is the first end-to-end invertible neural network to jointly model molecular property simulation such as $^{13}C$ NMR spectrum simulation and structure elucidation in the literature. By combining invertible architectures with conditional and latent-variable formulations, our model offers a principled alternative to purely discriminative or autoencoder-based approaches for spectrum–structure inference.

With respect to terminology, we use the term \textit{invertible neural network} here for a network which gives a bijective mapping between two information spaces. The term \textit{reversible neural network} is used in the literature for networks where the neurons have reversible functions, but where the network is not necessarily reversible as a whole. The purpose of those networks is mainly to enable memory savings by avoiding the need to save values in the nodes during training.

\section{Results and discussion}
\label{sec:results}

When training the model, we noticed that good results are achieved with a small number of training rounds. Figure~\ref{fig:training} and Table~\ref{tab:training} show that neither the loss nor the F1 value for $Y_{latent}$ improve beyond six epochs. We therefore have chosen five epochs for training for all subsequent evaluations. The increasing validation loss beyond epoch 5 whilst the training loss still improves suggests that overfitting occurs. Considering the small number of epochs, this could indicate that the amount of available data may be insufficient. We therefore select five epochs as an early-stopping point. Due to the limited size of the dataset and the rapid onset of overfitting, we restrict our evaluation to repeated training/validation splits (80\%/20\%). As a consequence, all reported performance metrics should be interpreted as validation performance rather than unbiased test estimates.

\begin{table}[ht]
\centering
\caption{Metrics of the model during 10 epochs of training.}
\label{tab:training}
\begin{tabular}{|l|l|l|l|l|l|}
\hline
& & \multicolumn{2}{c|}{Train} & \multicolumn{2}{c|}{Validation} \\
\textbf{Epoch} & \textbf{F1} & \textbf{Loss Y} & \textbf{Loss X} & \textbf{Loss Y} & \textbf{Loss X} \\ 
\hline
1 & 0.1053 & 0.1016 & 0.015 & 0.1079 & 0.0016\\ \hline
2 & 0.1381 & 0.1041 & 0.0014 & 0.1063 & 0.0015\\ \hline
3 & 0.1586 & 0.1013 & 0.0014 & 0.1065 & 0.0014\\ \hline
4 & 0.1719 & 0.1056 & 0.0015 & 0.1079 & 0.0016\\ \hline
5 & 0.1747 & 0.0806 & 0.0015 & 0.1145 & 0.0015\\ \hline
6 & 0.1762 & 0.0731 & 0.0015 & 0.1170 & 0.0015\\ \hline
7 & 0.1721 & 0.0725 & 0.0015 & 0.1244 & 0.0016\\ \hline
8 & 0.1731 & 0.0576 & 0.0015 & 0.1315 & 0.0016\\ \hline
9 & 0.1762 & 0.0528 & 0.0015 & 0.1436 & 0.0016\\ \hline
10 & 0.1707 & 0.0410 & 0.0015 & 0.1519 & 0.0016\\ \hline
\end{tabular}
\end{table}

\begin{figure}
    \centering
\begin{tikzpicture}
\begin{axis}[
    xlabel={Epoch},
    ylabel={Value},
     yticklabels={,,0.05,0.1,0.15},
    grid=major,
    legend pos=outer north east,
    legend style = {
        cells = {align = left}, 
        },          
    legend cell align = {left},     
]

\addplot[color=blue, mark=x] table {
    x y
    1 0.1053
    2 0.1381
    3 0.1586
    4 0.1719
    5 0.1747
    6 0.1762
    7 0.1721
    8 0.1731
    9 0.1762
    10 0.1707
};
\addlegendentry{F1}

\addplot[color=red, mark=x] table {
    x y
    1 0.1079
    2 0.1063
    3 0.1065
    4 0.1079
    5 0.1145
    6 0.1170
    7 0.1244
    8 0.1315
    9 0.1436
    10 0.1519
};
\addlegendentry{Loss Y\\Validation}

\addplot[color=green, mark=x] table {
    x y
    1 0.16
    2 0.15
    3 0.14
    4 0.16
    5 0.15
    6 0.15
    7 0.16
    8 0.16
    9 0.16
    10 0.16
};
\addlegendentry{Loss X (x 100)\\Validation}

\addplot[dashed, color=red, mark=x] table {
    x y
    1 0.1016
    2 0.1041
    3 0.1013
    4 0.1056
    5 0.0806
    6 0.0731
    7 0.0725
    8 0.0576
    9 0.0528
    10 0.0410
};
\addlegendentry{Loss Y\\Training}

\addplot[dashed, color=green, mark=x] table {
    x y
    1 0.15
    2 0.14
    3 0.14
    4 0.15
    5 0.15
    6 0.15
    7 0.15
    8 0.15
    9 0.15
    10 0.15
};
\addlegendentry{Loss X (x 100)\\Training}
\end{axis}
\end{tikzpicture}
\caption{F1 and loss values during training.}
\label{fig:training}
\end{figure}

For evaluating the performance of the model, a number of metrics are used. It should be noted that the absolute performance of the model is not as good as that of  specialized models. This is currently acceptable, since the invertible nature of the model is a novel concept.

\subsection{The network can predict spectra}

The model predicts spectra as a sequence of 128 0s and 1s. A typical measure for the correctness of such a task is the F1-value, calculated on $Y_{latent}$. Running our model 10 times with different, random train/validation splits gives F1 values between 0.1548 and 0.1839, with a mean of 0.1695. Clearly, this is not a very good result as such, but it shows that the model learns what it is intended to learn. Given the high sparsity of the output vectors (on average 6.43\% 1s per spectrum), a random predictor with the same probability of 1s and 0s achieves an expected F1 of about 0.0643, indicating that the obtained values are substantially above chance. In our context, the number and relative positions of 1s are more important than exact matches at all positions. In Table~\ref{tab:spectrapredicted} we therefore show a few random examples of real spectra and the associated predictions, clearly showing that they are similar. The 127, which occurs several times in the predicted spectra, is the value for a shift greater than or equal 204.6. This suggests that the model has learned to associate high shift values with a distinct output bin, although it still activates this bin in incorrect contexts. 

\begin{table}[ht]
\centering
\caption{Some typical example of real and predicted spectra of the invertible model. The numbers give the position of the 1s if the spectra are represented as 128 bit binned spectra.}
\label{tab:spectrapredicted}
\begin{tabular}{|l|l|l|}
\hline
\textbf{nmrshiftdb2} & \textbf{Real} & \textbf{Predicted}\\
\textbf{molecule ID} & {} & {}\\
\hline
20181820 & 14, 20, 22, 78, 81 & 11, 13, 85, 127\\
20097935 & 13, 77, 79, 80, 86, 88 &76, 77, 78, 79, 80, 82, 83, 84\\
60033845  & 9, 11, 13, 17, 18, 22, 23, & 10, 12, 13, 15, 17, 18, 19, 20, 21, \\
& 24, 33, 74, 84, 85, 92  & 22, 23, 24, 27, 28, 75, 77, 83, 86, 88\\
10016372  & 3, 12 &  8, 14, 16, 20\\
60063671  & 8, 10, 13, 19, 47, 49, 67, 68, 80, 87, 89 &  8, 11, 40, 41, 46, 47, 49, 50,\\
 & &  68, 80, 82, 86, 89 \\
60033894  & 7,  8, 12, 14, 16, 18, 21, 22, 80, 82 &  7,  8, 12, 14, 17, 18, 19, 20, 21, \\
 & & 23, 71, 80\\
10023267  & 8, 10, 27, 74, 76, 77, 78, 85, 86, 90, 93 & 12, 15, 23, 74, 77, 78, 80, 89, 90, 92\\
10005628  & 7,  8, 41, 53 &  46,  72, 127\\
10016436 & 10, 71, 80, 83, 86 &  11,  46,  72,  82,  83,  85, 127\\
10016490 & 14, 75, 81, 89, 93 &  85, 127\\
\hline
\end{tabular}
\end{table}

\subsection{The network is invertible for trained values}

For an invertible network, as a minimum requirement, it should be possible to reconstruct the input from any output in the training set. In our case, this is the possibility to reconstruct the input from the complete output of $Y_{latent}$ and $Z_{free}$. In theory, this should be perfect. In practice, rounding errors in the network give rise to negligible numerical errors. In our case, the maximum values observed during ten runs are in the range from 0.00009 to 0.00015 and the mean is between 0.00005 and 0.00007. These numbers indicate perfect reconstruction, as expected from an invertible model.

\subsection{$Z_{free}$ primarily captures residual variation}

Since all meaningful content on the output side of the network is in $Y_{latent}$, we want $Z_{free}$ to be independent from $Y_{latent}$ and to be well-behaved. Table~\ref{tab:zfree} shows that this is supported by the actual values. While the mean of $Z_{free}$ is close to zero, its standard deviation is substantially below one, indicating that the learned distribution is more concentrated than the ideal standard normal. This suggests that the regularization of $Z_{free}$ is only partially effective.

\begin{table}[ht]
\centering
\caption{Real and ideal values for $Z_{free}$ statistics.}
\label{tab:zfree}
\begin{tabular}{|l|l|l|l|l|}
\hline
& \multicolumn{4}{c|}{\textbf{Over validation set}} \\
\textbf{Metric} & \multicolumn{2}{c|}{\textbf{Real}} & \multicolumn{2}{c|}{\textbf{Ideal}} \\ 
\textbf{per Sample} & \textbf{Mean} & \textbf{Std} & \textbf{Mean} & \textbf{Std}\\ 
\hline
$Z_{free}$ mean & 0.0858 & 0.0334 & ~0 & small\\
$Z_{free}$ std & 0.5861	& 0.0795 & ~1 & small\\
\hline
\end{tabular}
\end{table}

\subsection{The data manifold is well structured}

We study how stable the inverse mapping is when we change the free latent part $Z_{free}$ while keeping the spectrum code $Y_{latent}$ fixed. We report two variants, because $Z_{free}$ can be taken either from the forward pass of a real sample or from the learned prior.

\textbf{Local perturbations (around validation samples).}
For each validation input $x_i$, we compute its latent split $(y_i, z_i)=f(x_i)$, where $y_i$ are the $Y_{latent}$ coordinates and $z_i$ are the $Z_{free}$ coordinates. We then add Gaussian noise to $z_i$ and measure how much the reconstructed input changes:
\begin{equation}
\label{eq:conditional_variation_local}
\mathrm{CD}_{\mathrm{local}}(\varepsilon)
=
\mathbb{E}_{i \in \mathrm{val}}
\mathbb{E}_{\xi \sim \mathcal{N}(0, I_d)}
\Big[
\big\| f^{-1}(z_i + \varepsilon \xi, y_i) - f^{-1}(z_i, y_i) \big\|_1
\Big],
\end{equation}
where $d=\dim(Z_{free})$ and $\|\cdot\|_1$ is the $\ell_1$ norm over the flattened reconstructed input tensor $X$.

\textbf{Prior perturbations (around typical $Z_{free}$ samples).}
To quantify sensitivity for typical draws of the free latent, we also sample $z \sim p(z)$ from the learned prior and repeat the same perturbation while keeping $y_i$ fixed:
\begin{equation}
\label{eq:conditional_variation_prior}
\mathrm{CD}_{\mathrm{prior}}(\varepsilon)
=
\mathbb{E}_{i \in \mathrm{val}}
\mathbb{E}_{z \sim p(z)}
\mathbb{E}_{\xi \sim \mathcal{N}(0, I_d)}
\Big[
\big\| f^{-1}(z + \varepsilon \xi, y_i) - f^{-1}(z, y_i) \big\|_1
\Big].
\end{equation}

Finally, we normalize by the typical magnitude of $X$ to obtain a relative deviation:
\begin{equation}
\mathrm{rCD}_{\star}(\varepsilon)=
\frac{\mathrm{CD}_{\star}(\varepsilon)}
{\mathbb{E}_{i\in\mathrm{val}}\big\|f^{-1}(z_i,y_i)\big\|_1},
\qquad \star \in \{\mathrm{local},\mathrm{prior}\}.
\end{equation}

\begin{table}[ht]
\centering
\caption{Conditional deviation under perturbations of $Z_{free}$ at fixed $Y_{latent}$.}
\label{tab:manifold}
\begin{tabular}{|l|l|l|}
\hline
\textbf{Radius $\epsilon$} & \textbf{$\mathrm{CD}_{\mathrm{local}}(\epsilon)$} & \textbf{$\mathrm{rCD}_{\mathrm{local}}(\epsilon)$} \\
\hline
0.01 & 0.598314 & 0.140901 \\
0.05 & 2.760628 & 0.635761 \\
0.10 & 5.379945 & 1.106703 \\
0.20 & 10.559302 & 1.083014 \\
0.50 & 21.109917 & 0.578593 \\
1.00 & 38.130421 & 0.598101 \\
\hline
\end{tabular}
\end{table}

The conditional deviation increases approximately linearly for small $\epsilon$, indicating local sensitivity to perturbations in $Z_{free}$. Importantly, the relative deviation does not grow beyond $\epsilon=0.1$, suggesting that the inverse map remains locally stable over the tested perturbation range.
To look beyond the random sample, we have calculated the conditional variation at fixed $Y_{latent}$ across a given set of $Z_{free}$ samples, where $Z_{free}$ comes from the learned prior. Here, we get a mean conditional variation of 3.4648 and a standard deviation of 0.8280. This indicates that changes in $Z_{free}$ lead to meaningful variation in the reconstructed input even when $Y_{latent}$ is fixed. This behavior is consistent with the intended role of $Z_{free}$ as a carrier of residual variation beyond the spectrum code.

\subsection{Generated X values are meaningful}

Finally, we can show that the X values, representing structures, generated by inverting the network from spectra from the validation set, which were not used during training, combined with a random $Z_{free}$, are meaningful. It should be clear that the results can hardly be an exact result, since for this, a very specific arrangement of only a few 1s in the matrices is required. During training, there is nothing that really enforces those arrangements. We have also seen that $Y_{latent}$ (our actual result) is not as independent from $Z_{free}$ as it could be, meaning the random selection of $Z_{free}$ still has an influence on the result of the inversion.

Because of this, we have two tests here, which are rather coarse, but still show that the model has learned something about even those cases. First, we have calculated the correlation coefficient between the number of 1s in the real inputs and the reconstructed ones. For ten rounds, the mean correlation coefficient was 0.1863, with a minimum of 0.0971 and a maximum of 0.2284. Second, we have calculated the correlation between the number of 1s in the fourth array of the input for each molecule, which contains the aromatic bonds. Because aromatic atoms are typically represented in the NMR spectrum by high shift values, a particularly strong connection could be expected here. The correlation coefficient was 0.1775 (min=0.09563, max=0.2073) for ten runs.

While these correlations are weak, they are consistently positive across runs and exceed a purely random result, which would have correlation 0, indicating that the model preserves coarse structural information. It clearly shows that the inversion in this case only works very roughly. Still, the network is invertible for some cases and keeps some connection between input and output even in other, more difficult cases.

\section{Methods}

\subsection{Data}

The data used in this work are from nmrshiftdb2 \cite{RN2165,RN1117}. We have selected all structures which contain only carbon and hydrogen atoms, which have up to 17 carbon atoms, and for which a one-dimensional $^{13}C$ spectrum exists. There were 1567 such structures. Those structures and their assigned spectra have been encoded into sequences of 0s and 1s, both simplified compared to the original data.

For the structures, we encode the bonds between carbon atoms as an adjacency matrix, which holds the numbers 1, 2, and 3 for single, double, and triple bonds. In case of aromatic bonds, the numbers are incremented by 5. So, aromaticity is recorded in addition to the single and double bonds in the alternating single/double bond depiction. The atoms are not encoded as such, but are contained in the adjacency matrix. This means that other properties, e.g. charges, are not (directly) encoded. Since molecules with only carbon and hydrogen atoms have been selected, and the hydrogen atoms are disregarded, there is no need to encode the element. Since the adjacency matrix always has zeros on the diagonal (there are no bonds from an atom to itself) and is symmetric (bonds are symmetric), the number of significant digits for each molecule is 1+...+16=136. In the reversibledata.csv file, those 136 numbers are in the C column. Atom ordering is taken directly from how structures are saved in nmrshiftdb2 and kept fixed; no attempt is made to enforce permutation invariance.

The spectrum has been converted into 1024 bins of width 0.2 ppm, where the first bin represents anything less than zero, and the last anything greater than or equal 204.6. A bin is set to 1, if there are any number of peaks in the range. Those 1024 0/1s are in the D column of the reversibledata.csv file. Columns A and B give the molecule and spectrum ID in nmrshiftdb2 used.

\subsection{The network architecture}

The invertible neural network we have been using is based on the i-RevNet architecture \cite{jacobsen2018irevnet,DBLP:journals/corr/abs-1802-07088}, but differs in how invertibility is maintained at the output stage. It should be noted that the architecture presented by Jacobsen \textit{et al.} is not fully invertible, in particular, the final classification step in the bijective network is not invertible. The RevNet block used as the main component, however, is fully invertible. Using this block, we construct a conditional invertible neural network (cINN, \cite{Ardizzone2021cINN}) which is fully invertible.

At the core of our network is a block which increases the number of channels by a factor of four and halves the x and y dimensions of the input layer. This block follows the implementation in i-RevNet. In this way, a multi-dimensional input can be transformed to a one-dimensional output in a provably fully invertible fashion. In our case, the input is structured as a 4x16x16 array, where 4 is the initial number of channels. Each of the matrices holds a 1 for a bond, and a 0 otherwise. The first matrix has single bonds, the second double bonds, and the third triple bonds. The fourth matrix indicates if a bond is an aromatic bond. This is in addition to the other matrices, so an aromatic ring is recoded as a number of single and double bonds with an aromatic property recorded in addition. Since atoms cannot bond to themselves, a 16x16 matrix can hold the information for the complete molecule. We do not include the symmetric bonds, so a part of the matrices is always empty, which means set to 0. The network produces an output of 1024 channels with a 1x1 matrix for each. In practice, this is a 1024 bit vector. The overall network architecture is shown in Figure~\ref{fig:architecture}.

\begin{figure}
    \centering
\hspace*{.2cm}\begin{tikzpicture}[x=0.2cm,y=0.1cm]
\newcommand{\slantedgrid}[4]{%
   \pgfmathtruncatemacro{\result}{#1+#3}
   \foreach \x in {#1,...,\result} \draw (\x,#2) -- ++(#4,#4);%
   \pgfmathtruncatemacro{\result}{#2+#4}
   \foreach \y in {#2,...,\result} \draw (#1+\y-#2,\y) -- ++(#3,0);%
 }
\fill[black,fill] (0,0) to (1,0)
    to (2,1)
    to (1,1)
    to cycle;
\fill[black,fill] (4,2) to (5,2)
    to (6,3)
    to (5,3)
    to cycle;
\fill[black,fill] (5,0) to (6,0)
    to (7,1)
    to (6,1)
    to cycle;
\fill[black,fill] (2,1) to (3,1)
    to (4,2)
    to (3,2)
    to cycle;
\fill[black,fill] (6,3) to (7,3)
    to (8,4)
    to (7,4)
    to cycle;
\slantedgrid{0}{0}{16}{16}
\fill[white,fill] (0,3.5) to (16.2,3.5)
    to (32.5,20)
    to (16,20)
    to cycle;
\slantedgrid{0}{4}{16}{16}
\fill[white,fill] (0,7.5) to (16.2,7.5)
    to (32.5,24)
    to (16,24)
    to cycle;
\fill[black,fill] (2,9) to (3,9)
    to (4,10)
    to (3,10)
    to cycle;
\fill[black,fill] (6,11) to (7,11)
    to (8,12)
    to (7,12)
    to cycle;
\fill[black,fill] (5,8) to (6,8)
    to (7,9)
    to (6,9)
    to cycle;
\slantedgrid{0}{8}{16}{16}
\fill[white,fill] (0,11.5) to (16.2,11.5)
    to (32.5,28)
    to (16,28)
    to cycle;
\fill[black,fill] (0,12) to (1,12)
    to (2,13)
    to (1,13)
    to cycle;
\fill[black,fill] (4,14) to (5,14)
    to (6,15)
    to (5,15)
    to cycle;
\fill[black,fill] (8,16) to (9,16)
    to (10,17)
    to (9,17)
    to cycle;
\slantedgrid{0}{12}{16}{16}

\path (33,16) -| node[coordinate] (n1) {} (33,28);
\draw[thick,decorate,decoration={brace,amplitude=3pt}]
            (33,28) -- (n1) node[midway, right=4pt] {4};
\node[] at (8, -1.7)    {16};
\node[rotate=35] at (25.5, 7.5)    {16};
\end{tikzpicture}

\begin{tikzpicture}
\node[] (A) at (1,0) {};
\node[] (B) at (1,1) {};
\node[] (C) at (1.3,0) {};
\node[] (D) at (1.3,1) {};
\draw[->] (A) -- (B);
\draw[<-] (C) -- node[rotate=0,right] {invertible transformation} (D);
\end{tikzpicture}

\hspace*{-1.2cm}\begin{tikzpicture}[x=0.2cm,y=0.1cm]
\newcommand{\slantedgrid}[4]{%
   \pgfmathtruncatemacro{\result}{#1+#3}
   \foreach \x in {#1,...,\result} \draw (\x,#2) -- ++(#4,#4);%
   \pgfmathtruncatemacro{\result}{#2+#4}
   \foreach \y in {#2,...,\result} \draw (#1+\y-#2,\y) -- ++(#3,0);%
 }
\slantedgrid{0}{0}{8}{8}
\fill[white,fill] (0,1.5) to (8.2,1.5)
    to (16.5,10)
    to (8,10)
    to cycle;
\fill[black,fill] (4,2) to (5,2)
    to (6,3)
    to (5,3)
    to cycle;
\slantedgrid{0}{2}{8}{8}
\fill[white,fill] (0,3.5) to (8.2,3.5)
    to (16.5,12)
    to (8,20)
    to cycle;
\slantedgrid{0}{4}{8}{8}
\fill[white,fill] (0,5.5) to (8.2,5.5)
    to (16.5,14)
    to (8,14)
    to cycle;
\filldraw[black] (3,6.5) circle (2pt);
\filldraw[black] (4,6.5) circle (2pt);
\filldraw[black] (5,6.5) circle (2pt);
\fill[black,fill] (5,12) to (6,12)
    to (7,13)
    to (6,13)
    to cycle;
\slantedgrid{0}{8}{8}{8}
\path (17,8) -| node[coordinate] (n1) {} (17,16);
\draw[thick,decorate,decoration={brace,amplitude=3pt}]
            (17,16) -- (n1) node[midway, right=4pt] {16};
\node[] at (4, -1.7)    {8};
\node[rotate=35] at (14, 4)    {8};
\end{tikzpicture}

\begin{tikzpicture}
\node[] (A) at (1,0) {};
\node[] (B) at (1,1) {};
\node[] (C) at (1.3,0) {};
\node[] (D) at (1.3,1) {};
\draw[->] (A) -- (B);
\draw[<-] (C) -- node[rotate=0,right] {invertible transformation} (D);
\end{tikzpicture}

\hspace*{-2.0cm}\begin{tikzpicture}[x=0.2cm,y=0.1cm]
\newcommand{\slantedgrid}[4]{%
   \pgfmathtruncatemacro{\result}{#1+#3}
   \foreach \x in {#1,...,\result} \draw (\x,#2) -- ++(#4,#4);%
   \pgfmathtruncatemacro{\result}{#2+#4}
   \foreach \y in {#2,...,\result} \draw (#1+\y-#2,\y) -- ++(#3,0);%
 }
\slantedgrid{0}{0}{4}{4}
\fill[white,fill] (0,1.5) to (4.2,1.5)
    to (8.5,6)
    to (4,6)
    to cycle;
\slantedgrid{0}{2}{4}{4}
\fill[white,fill] (0,3.5) to (4.2,3.5)
    to (8.5,8)
    to (4,8)
    to cycle;
\slantedgrid{0}{4}{4}{4}
\fill[white,fill] (0,5.5) to (4.2,5.5)
    to (8.5,10)
    to (4,10)
    to cycle;
\filldraw[black] (2,6.5) circle (2pt);
\filldraw[black] (3,6.5) circle (2pt);
\filldraw[black] (4,6.5) circle (2pt);
\fill[black,fill] (2,10) to (3,10)
    to (4,11)
    to (3,11)
    to cycle;
\slantedgrid{0}{8}{4}{4}
\path (9,4) -| node[coordinate] (n1) {} (9,12);
\draw[thick,decorate,decoration={brace,amplitude=3pt}]
            (9,12) -- (n1) node[midway, right=4pt] {64};
\node[] at (2, -1.7)    {4};
\node[rotate=35] at (7, 1)    {4};
\end{tikzpicture}

\begin{tikzpicture}
\node[] (A) at (1,0) {};
\node[] (B) at (1,1) {};
\node[] (C) at (1.3,0) {};
\node[] (D) at (1.3,1) {};
\draw[->] (A) -- (B);
\draw[<-] (C) -- node[rotate=0,right] {invertible transformation} (D);
\end{tikzpicture}

\hspace*{-2.5cm}\begin{tikzpicture}[x=0.2cm,y=0.1cm]
\newcommand{\slantedgrid}[4]{%
   \pgfmathtruncatemacro{\result}{#1+#3}
   \foreach \x in {#1,...,\result} \draw (\x,#2) -- ++(#4,#4);%
   \pgfmathtruncatemacro{\result}{#2+#4}
   \foreach \y in {#2,...,\result} \draw (#1+\y-#2,\y) -- ++(#3,0);%
 }
\slantedgrid{0}{0}{2}{2}
\fill[white,fill] (0,1.5) to (2.2,1.5)
    to (4.5,4)
    to (2,4)
    to cycle;
\slantedgrid{0}{2}{2}{2}
\fill[black,fill] (1,3) to (2,3)
    to (3,4)
    to (2,4)
    to cycle;
\fill[white,fill] (0,3.5) to (2.2,3.5)
    to (4.5,6)
    to (2,6)
    to cycle;
\slantedgrid{0}{4}{2}{2}
\fill[white,fill] (0,3.5) to (2.2,5.5)
    to (4.5,8)
    to (4,8)
    to cycle;
\filldraw[black] (1,6.5) circle (2pt);
\filldraw[black] (2,6.5) circle (2pt);
\filldraw[black] (3,6.5) circle (2pt);
\slantedgrid{0}{8}{2}{2}
\path (5,0) -| node[coordinate] (n1) {} (5,10);
\draw[thick,decorate,decoration={brace,amplitude=3pt}]
            (5,10) -- (n1) node[midway, right=4pt] {256};
\node[] at (2, -1.7)    {2};
\node[rotate=35] at (4, 0)    {2};
\end{tikzpicture}

\begin{tikzpicture}
\node[] (A) at (1,0) {};
\node[] (B) at (1,1) {};
\node[] (C) at (1.3,0) {};
\node[] (D) at (1.3,1) {};
\draw[->] (A) -- (B);
\draw[<-] (C) -- node[rotate=0,right] {invertible transformation} (D);
\end{tikzpicture}

\hspace*{1.6cm}\begin{tikzpicture}[x=0.2cm,y=0.1cm]
\newcommand{\slantedgrid}[4]{%
   \pgfmathtruncatemacro{\result}{#1+#3}
   \foreach \x in {#1,...,\result} \draw (\x,#2) -- ++(#4,#4);%
   \pgfmathtruncatemacro{\result}{#2+#4}
   \foreach \y in {#2,...,\result} \draw (#1+\y-#2,\y) -- ++(#3,0);%
 }
\slantedgrid{0}{0}{1}{1}
\slantedgrid{0}{2}{1}{1}
\fill[black,fill] (0,2) to (1,2)
    to (2,3)
    to (1,3)
    to cycle;
\slantedgrid{0}{4}{1}{1}
\filldraw[black] (0,6.5) circle (2pt);
\filldraw[black] (1,6.5) circle (2pt);
\filldraw[black] (2,6.5) circle (2pt);
\slantedgrid{0}{8}{1}{1}
\path (3,1) -| node[coordinate] (n1) {} (3,9);
\draw[thick,decorate,decoration={brace,amplitude=3pt}]
            (3,9) -- (n1) node[midway, right=4pt] {1024};
\node[] at (1, -1.7)    {1};
\node[rotate=35] at (2.5,-.5)    {1};
\node[] at (10,4) {=};
\draw (12,3) -- (28,3);
\draw (12,5) -- (28,5);
\draw (12,5) -- (12,3);
\draw (13,5) -- (13,3);
\fill[black,fill] (13,5) to (13,3)
    to (14,3)
    to (14,5)
    to cycle;
\draw (14,5) -- (14,3);
\draw (15,5) -- (15,3);
\filldraw[black] (16,4) circle (1pt);
\filldraw[black] (17,4) circle (1pt);
\filldraw[black] (18,4) circle (1pt);
\draw (19,5) -- (19,3);
\draw (20,5) -- (20,3);
\draw[dashed] (20,10) -- (20,0);
\node[] at (17.65, 10)    {conceptual split};
\draw (21,5) -- (21,3);
\draw (22,5) -- (22,3);
\filldraw[black] (23,4) circle (1pt);
\filldraw[black] (24,4) circle (1pt);
\filldraw[black] (25,4) circle (1pt);
\draw (26,5) -- (26,3);
\fill[black,fill] (26,5) to (26,3)
    to (27,3)
    to (27,5)
    to cycle;
\draw (27,5) -- (27,3);
\draw (28,5) -- (28,3);
\draw [decorate,decoration={brace,amplitude=5pt,mirror,raise=4ex}]
  (12,8) -- (20,8) node[midway,yshift=-3em]{128=$Y_{latent}$};
\draw [decorate,decoration={brace,amplitude=5pt,mirror,raise=4ex}]
  (20,8) -- (28,8) node[midway,yshift=-3em]{896=$Z_{free}$};
\end{tikzpicture}
\caption{The network consists of a sequence of invertible iRevNet blocks, which progressively transform the four input matrices into a 1D vector. The final latent space is conceptually partitioned into 128 bits representing a spectrum ($Y_{latent}$) and 896 unconstrained bits ($Z_{free}$). Black dots indicate omitted layers or cells, shown schematically for visual clarity. The black boxes indicate 1 values, where in the first layer a benzene molecule, with three single and three double bonds formint a ring, and all bonds marked as aromatic. 1s in further layers are put in randomly for illustrative purposes. }
\label{fig:architecture}
\end{figure}

Whilst at first sight this would mean that a direct prediction of the 1024 bit spectrum representation could be attempted, a reversible neural network cannot necessarily predict in both directions what is possible from the architecture. The important measure is entropy, and doing a rough entropy estimate for our input gives an entropy of 76 bits. We therefore have divided the output into 128 bits (to keep with the $2^x$ sizes) for the actual spectral features (referred to as $Y_{latent}$) and 1024-128=896 bits which absorb the remaining variation (referred to as $Z_{free}$). It should be noted that this is conceptual split we make on the result, the network still learns the full 1024 bits (but it is incentivized to get $Y_{latent}$ right by the loss function). A detailed examination of this is found in Section~\ref{sec:results}. For mapping from the 1024 bit spectrum to the 128 bit compression, a mapping where four bits are projected into one bit using an OR-function is used. This ensures that no information is lost and is equivalent to doing a coarser binning. Our model will only work on the 128 bit spectra. This is a limitation which stems from the requirement of an invertible network to have similarly sized information spaces on both sides.

The loss function is composed of four factors: For measuring the accuracy of the prediction, we have implemented a distance-aware Binary Cross-Entropy (BCE) loss function operating on $Y_{latent}$. The distance aware element is important since in the spectrum, which is represented by $Y_{latent}$, the position matters: For example, if a part of the real spectrum is encoded as 0100 then the prediction 0010 is better than 0001, although both do not have the 1 in the correct position. The second component penalizes if the inputs generated by reversing the network are outside the 0/1 range. The other components ensure the sparsity of the input matrices. So, the overall loss ensures good results in both directions.

\section{Future work}

Whilst we have shown in this work that invertible neural networks are a possible tool for spectrum prediction and structure elucidation, there is a significant amount of work to be done in the area. A first step is to gather more data from other sources than nmrshiftdb2 to better train and test the existing network. It should be possible to achieve ideal values in testing, where this is currently not the case. In a further step, larger structures and structures with heteroatoms can be included. At the same time, the spectra types used can also be extended. For this, work on the information state of structures and spectra is needed. It may be necessary to change or extend the network architecture with new (reversible) elements. The training process, including the loss function and the weights, is likely to need to be improved for these architectures to work.

\section{Conclusion}

We have presented a conditional invertible neural network, based on i-RevNet blocks, that is able to predict spectra from structures in one direction and elucidate structures from spectra in the other direction without explicit training for that. In order to maintain entropy on both sides of our network, we divide the output into useful information of 128 bits and leave the rest as residual noise. We demonstrate that the network can predict spectra and that it is invertible for the training examples. Furthermore, we demonstrate that the output outside of the 128 bits is mainly noise and that the data manifold is well structured. Finally, we have also shown that when used in reverse with unknown examples, the network still produces some chemically meaningful results.

\section*{Code availability}

The code for the invertible neural network is available from github at \url{https://github.com/stefhk3/RevMolSpec}.

\section*{AI declaration}

ChatGPT was used to analyse the i-RevNet code and to isolate functions for the code at \url{https://github.com/stefhk3/RevMolSpec}. No part of the paper was written by ChatGPT.

\bibliographystyle{splncs04}
\bibliography{achemso-demo}

\end{document}